\documentclass[preprint,journal]{vgtc}            

\onlineid{1032}

\ieeedoi{10.1109/TVCG.2024.3456143}

\vgtccategory{Research}

\vgtcpapertype{application/design study}

\title{Visualizing Temporal Topic Embeddings with a Compass}

\author{%
  \authororcid{Josiah S.\ Carberry}{0000-0002-1825-0097},
  Ed Grimley, and 
  Martha Stewart
}

\authorfooter{
  \begin{itemize}
  \item Daniel Palamarchuk, Lemara Williams, Brian Mayer, Chris North are with Virginia Polytechnic Institute and State University, VA.
        Email: {d4n1elp, lemaraw}@vt.edu, {bmayer, north}@cs.vt.edu
  \item Thomas Danielson and Larry Deschaine are at Savannah River National Laboratory.
        Email: Thomas.Danielson@srnl.doe.gov, larry.deschaine@srnl.doe.gov
  \item Rebecca Faust is at Tulane University.
        Email: rfaust1@tulane.edu
  \end{itemize}
  * Daniel Palamarchuk and Lemara Williams are co-first authors
}

\abstract{%
Dynamic topic modeling is useful at discovering the development and change in latent topics over time.
However, present methodology relies on algorithms that separate document and word representations.
This prevents the creation of a meaningful embedding space where changes in word usage and documents can be directly analyzed in a temporal context.
This paper proposes an expansion of the compass-aligned temporal Word2Vec methodology into dynamic topic modeling.
Such a method allows for the direct comparison of word and document embeddings across time in dynamic topics.
This enables the creation of visualizations that incorporate temporal word embeddings within the context of documents into topic visualizations.
In experiments against the current state-of-the-art, our proposed method demonstrates overall competitive performance in topic relevancy and diversity across temporal datasets of varying size.
Simultaneously, it provides insightful visualizations focused on temporal word embeddings while maintaining the insights provided by global topic evolution, advancing our understanding of how topics evolve over time.
}


\teaser{
  \centering
  \includegraphics[trim={0 8cm 0 8cm}, clip, width=\textwidth, alt={Teaser image of TimeLink sankey diagram used in the paper}]{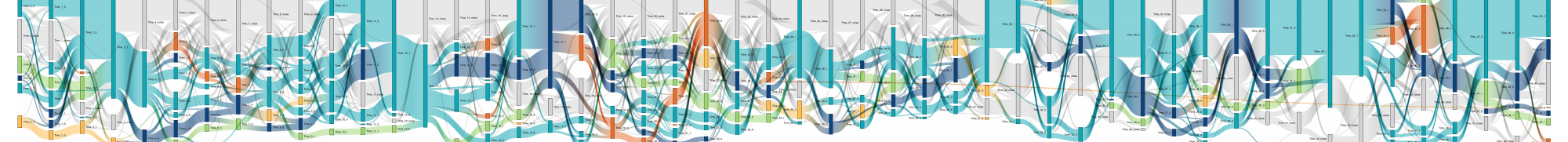}
  \caption{%
  	Temporal topics computed by TTEC and keyword evolution over time visualized with TimeLink.%
  }
  \label{fig:fullsankey}
}

\usepackage{amsmath,amsfonts}
\usepackage{algorithmic}
\usepackage{algorithm}
\usepackage{array}
\usepackage[caption=false,font=normalsize,labelfont=sf,textfont=sf]{subfig}
\usepackage{textcomp}
\usepackage{stfloats}
\usepackage{url}
\usepackage{verbatim}
\usepackage{graphicx}
\usepackage{cite}
\usepackage{graphicx}
\newcommand{\changed}[1]{\textcolor{black}{#1}}

\date{31 March, 2024}

\begin{document}

\title{Visualizing Temporal Topic Embeddings with a Compass}

\author{Daniel Palamarchuk*, Lemara Williams*, Brian Mayer, Thomas Danielson, Rebecca Faust, Larry Deschaine, Chris North
}

\markboth{Journal of \LaTeX\ Class Files,~Vol.~14, No.~8, August~2021}%
{Shell \MakeLowercase{\textit{et al.}}: A Sample Article Using IEEEtran.cls for IEEE Journals}


\maketitle


\newpage

\section{Introduction}


\changed{Historians, scholars, intelligence analysts and others often have large amounts of text documents that they must review, assess, synthesize, and gleam important events from.}
When analyzing a collection of documents that were generated over time,  understanding what topics exist and how these topics evolve and shift in definition 
over time provides insight into trends and behaviors in the domain described by the  documents~\cite{alexander2015task}.
\changed{For example, an intelligence analyst may need to know that an entity of interest has changed their social or business relationships.}
The temporal dynamics of topics help illustrate global trends of discussion in the corpus, such 
as what topics exist overall (i.e., a global topic structure), when topics are introduced, when topics stop being 
discussed, how the topics from more recent years differ from past years, and so on. 
Additionally, inspecting the movement of a given keyword (identified in the documents) 
between topics over time illustrates how the context and discussion around 
that keyword evolves as time progresses, which helps identify significant changes/events related to it.  

For example, consider \changed{an intelligence analyst} who seeks to identify significant events on a global scale related to nuclear energy by analyzing a corpus of millions of documents (e.g., news articles) collected over years, or possibly decades.
\changed{An analyst} can expect that when a significant event occurs, the context (i.e., within documents added to the corpus around the time of the event) surrounding a specific entity or event (e.g., a specific nuclear power plant) will temporarily shift, causing a change in the associated topic - discussing the event as it unfolds, before shifting back toward a more general discussion or to an evolving topic space. Thus, to assist in identifying significant events, the \changed{analyst} can leverage the dynamic topic structure of the embedding space to understand how the discussion surrounding different aspects of nuclear energy shifts and evolves over time. In other words, the \changed{analyst} can evaluate how topics associated with specific documents and keywords/phrases evolve as information is added to the text corpus, thereby highlighting detailed evidence from relevant documents as shifts occur. 

\changed{Examining millions of documents by hand to organize them into topics is infeasible. Users can apply current dynamic topic modeling practices, such as BERTopic~\cite{grootendorst22bertopic} or sequential Latent Dirichlet Allocation~\cite{blei06dynamic}, to identify topics, mapping documents to identified topics. However, to understand the dynamics of even a single topic, users must still investigate large quantities of documents and manually build a temporal illustration of the topic's evolution. Furthermore, to examine how the context of keywords of interest evolve over time
additional analysis is needed to identify which topics contain those keywords and how they change over time. Thus, there is a clear need for visual analytics (VA) to enable users to explore and understand the temporal dynamics in a large corpus of documents. 
}

\changed{To fully support temporal analysis of large corpora of documents, a VA system must empower users to interactively inspect how both the topics space and the composition of topics changes over time, relative to both documents and keywords, as well as how the semantics of a given keyword evolve over time. Additionally, the VA system must provide clear visual cues that highlight significant changes in topic composition and keyword behavior, as users in an exploratory setting will not know where to look for significant events.} To support topic-centric tasks involving both documents and keywords, the visual system must
(1) capture and illustrate the evolution of the topics in a collection of documents across time and
(2) embed keywords in the same topic embedding space as the documents to demonstrate how and why they shift across topics over time.
While research exists that enables visual text analytics through topic modeling, prior methods lack support for inspecting both the evolution of topics and the movement of keywords across topics.
\changed{Additionally, existing Dynamic Topic Modeling (DTM) approaches, such as sequential LDA and BERTopic, do not support integrating data from different embedding spaces, such as from both documents and keywords, into the same dynamic topic space. }

\changed{Thus, to create such a VA system, two questions must be addressed: }
\begin{itemize}    
    \item How can a single embedding space that combines words, documents, and topics from time-varying text data be created?
     \item How can a VA system interactively visualize this combined embedding space to reveal the temporal dynamics of a corpus?
\end{itemize}

To address these questions, we propose a \changed{VA system, TimeLink, backed by a new method for DTM -  Temporal Topic Embeddings with a Compass (TTEC), } 
to illustrate the evolving topic space. \changed{TTEC, building upon a prior methodology for compass-aligned temporal embeddings~\cite{carlo19},}
creates a global embedding space from the entire collection of words and documents and local embedding spaces for each slice in time. Thus, with TTEC, we create and couple global topics with the local, time-specific embeddings of keywords and documents, to capture the changing context over time in a hierarchical way. \changed{TimeLink is then able to visualizes these} global topics and local embeddings, to illustrate how the global topics evolve and how keywords shift across topics over time (i.e., as embeddings and word usage change over time).
Our contributions are as follows:
\begin{itemize}
    \item An expansion of compass-aligned temporal Word2Vec methodology into DTM including word and document embeddings
    \item Visualization for temporal word embeddings within the context of global topic evolution
    \item Two case studies with domain experts demonstrating the proposed method's ability to capture and illustrate the evolution of topics driven by underlying events 
    \item A quantitative comparison of TTEC with other DTM methods
\end{itemize}

\section{Related Work}
\label{sec:rw}

\subsection{Temporal Word Embeddings}


Word2vec \cite{mikolov13efficient} is a family of architectures built off of the distributional hypothesis of words \cite{harris54distributional}, which holds that a word can be defined by the words that surround it.
Similar words will be found in similar surroundings, and the Word2vec algorithm attempts to algorithmically capture that hypothesis using a shallow neural network to generate a word vector embedding representation using a surrounding context.
This architecture has been expanded to incorporate document and tag embeddings \cite{le14} to allow a comparison of words to documents.

Hamilton et al. extended Word2Vec to determine how the contexts that words appear in change over time~\cite{hamilton18}
They separate a corpus into time slices and learn context within each time slice.
Since each time slice possesses a different context, embedding words in individual time slices results in different embeddings that are unique to that time slice.
However, for these slices to be comparable, they need to establish a notion of slice-wise alignment, since otherwise it would be difficult to compare word movements (context changing) between time slices.
Several methods exist to align word embeddings.
Future time slices can be initialized from their previous one, or by applying a linear transformation to ensure pairwise similarity \cite{hamilton18,szymanski17temporal}.
The joint approach creates vector similarity through either global vectors or a smoothing process performed during training\cite{yao18,rudolph18dynamic}.

Compass-Aligned Distributed Embeddings (CADE) \cite{carlo19,bianchi20} are an alignment method that does not rely on a global vector or any notion of obtaining current aligned word embeddings from previous ones.
Instead, CADE initially trains an atemporal embedding model using the entire corpus to generate a shallow neural network, called a ``compass.''
It then keeps the hidden layer, known as the ``target embedding space,'' for the training of the time slices, where it remains unchanged.
For each time slice, CADE creates a different network, initializing the hidden layer with the weights in the compass and training each network using the local time slice corpus. 
Since the hidden embedding is frozen, obtaining similar outcomes from the neural network across time slices would require similar inputs, thus creating alignment across time slices.
Although CADE is the general term, the current method is focused on obtaining aligned word embeddings \cite{carlo19}, which are called Temporal Word Embeddings with a Compass (TWEC).
This paper will propose two additions to the CADE family of algorithms.

\subsection{Topic Modeling}

Topic modeling attempts to discern topics from a set of documents.
This can be done with Latent Dirichlet Allocation (LDA) \cite{blei03latent}, which classifies documents as bags of words composed of a combination of topics.
LDA was expanded upon to include Word2vec to describe the topics in the Embedded Topic Model (ETM) \cite{dieng20topic}.
Recently, topic modeling techniques have adopted the use of separate algorithms to create a topic space.
These algorithms commonly use UMAP to create a sensibly clustered low-dimensional space and HDBSCAN to discover and classify these clustered spaces.
Top2vec \cite{angelov20} creates a topic space out of the Doc2vec embeddings using UMAP and HDBSCAN.
The topics in these spaces are then described using the Word2vec embeddings that are closest to the centroid of the topics.
BERTopic \cite{grootendorst22bertopic} uses embeddings generated by Sentence-BERT \cite{reimers19sentencebert} and adds a class-based TF-IDF to generate topic descriptors in the absence of a way to generate word embeddings that are comparable to document embeddings.

Dynamic topic modeling (DTM) takes the topic modeling approach and expands it temporally to allow a user to view the evolution of a topic over time.
This was first thought of as a sequential LDA \cite{blei06dynamic} that attempted to smoothly create topic representations in time slices that take into consideration word and topic distribution in previous time slices.
ETM was similarly expanded to the temporal dimension with Dynamic ETM \cite{dieng19dynamic}.
BERTopic expands to DTM \cite{grootendorst22bertopic} by separating documents into time slices after putting them into topics.
This ensures that similar documents across different time slices maintain the same topic.
c-TF-IDF can then be performed on topics per time slice .

\begin{figure}[t]
    \includegraphics[trim={0 3.9cm 0 0},clip,width=\linewidth, alt={Heatmap of temporal cosine similarities between terms related to nuclear energy.}]{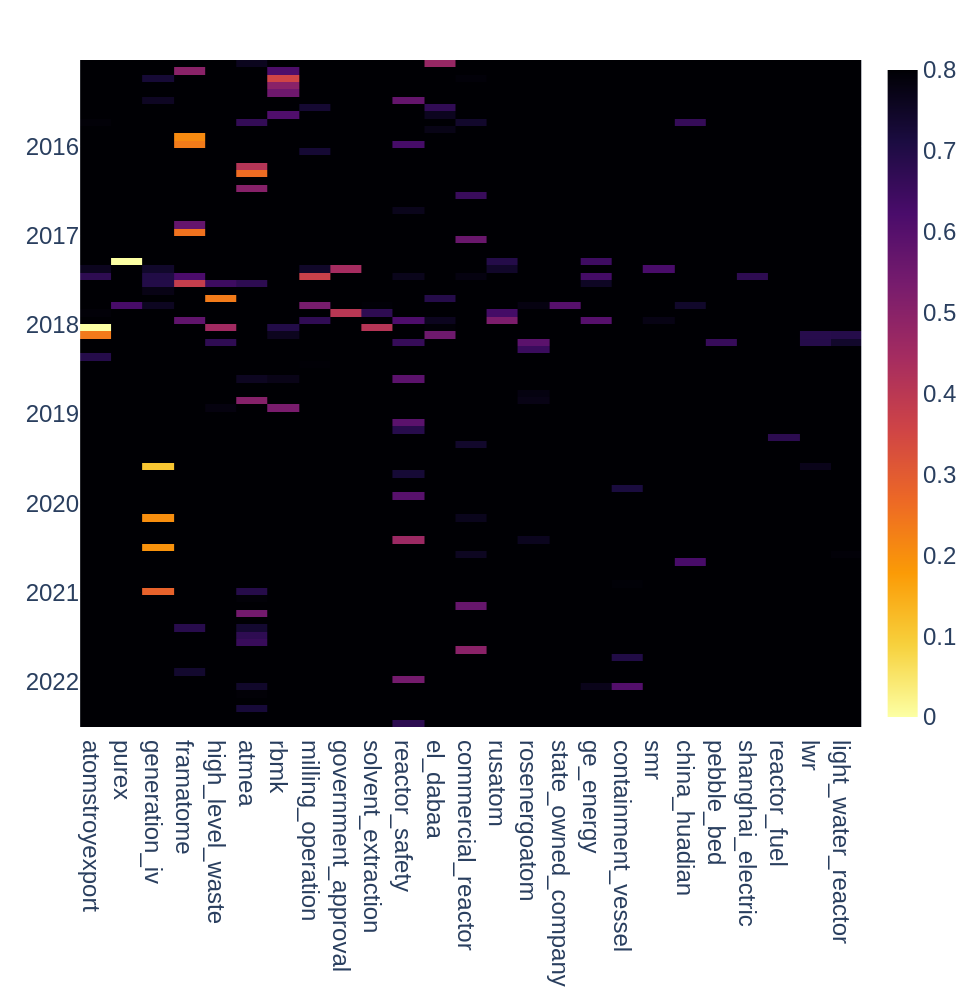}
    \caption{Heatmap represents total change in cosine similarities for each term for temporal word embeddings made using TWEC.}
    \label{fig:base_heatmap}
    \vspace{-2em}
\end{figure}

\subsection{Visualization of Topic Patterns and Evolution}
Many methods exist for visualizing the global topic space of a document corpus~\cite{sievert2014ldavis,lee2012ivisclustering}
Some works aim at visualizing topics across multiple corpora (akin to time slices) 
~\cite{wang2016ideas,wang2016topicpanorama}.
However, the most relevant works to the methods proposed here are those that visualize topics in dynamic spaces to illustrate patterns in, and evolutions of, topics. 

Several methods exist to visualize topic spaces over time. 
Mei et al. create a theme graph to illustrate similar themes (nodes) across time (via edges)~\cite{mei2005discovering}. 
Many works take a stacked area graph approach to show how topics evolve over time~\cite{sun2014evoriver,xu2013visual,liu2012tiara,gunnemann2013interactive,havre2002themeriver}. ThemeRiver introduces streams to illustrate themes (keywords) and show how they change in strength over time. Subsequent methods build on this metaphor in various ways, such as creating streams of topics built from multiple keywords~\cite{liu2012tiara, gunnemann2013interactive}, using topic trees to identify topics ~\cite{dou2013hierarchicaltopics}, and illustrating the competition and cooperation of topics in social media~\cite{xu2013visual,sun2014evoriver}. 

Where previous stacked area approaches show the trends of fixed global topics, several other methods visualize the dynamics of splitting and merging topics over time. TextFlow creates streams for topics that split and merge as topics evolve~\cite{cui2011textflow}.
Cui et al. expand TextFlow to visualize evolving hierarchical topic trees~\cite{cui2014hierarchical}.
Jiang and Zhang employ Sankey diagrams to illustrate the topic evolution and correlation across three research domains~\cite{jiang2016text}.
\changed{TopicFlow~\cite{malik13topicflow} uses aligned LDA topics as a backend for its own Sankey Diagram, which is color-coded based on whether it is an emerging or continuing topic.}
Gad et al. introduce ThemeDelta, which segments ThemeRivers to identify key breakpoints in theme evolution and groups theme streams by similarity to illustrate the evolution of global topics~\cite{gad2015themedelta}.
ThemeDelta presents the method most similar to the visualization proposed here.
However, ThemeDelta visualizes each individual theme as a stream and creates topics ad-hoc based on the similarity of themes at each time point.
In contrast, our approach identifies global topics and illustrates how clusters of keywords merge and split to define these topics over time.






\section{Baseline}
\label{sec:baseline}

\subsection{Nuclear Corpus Dataset}
\label{sub:nuclear}
Examples in this paper were generated using the ``Nuclear Corpus'' dataset.
It was gathered using a keyword search for articles with terms related to domestic and foreign nuclear energy policy from NewsAPI. 
The dataset spans between January 2015 and July 2022 and contains 5,637,381 articles with a total of 28,663,811,702 words after pre-processing, which involves the removal of stopwords.

\subsection{Temporal word embedding visualizations}

\begin{figure}[t]
    \centering
    \includegraphics[width=0.5\textwidth]{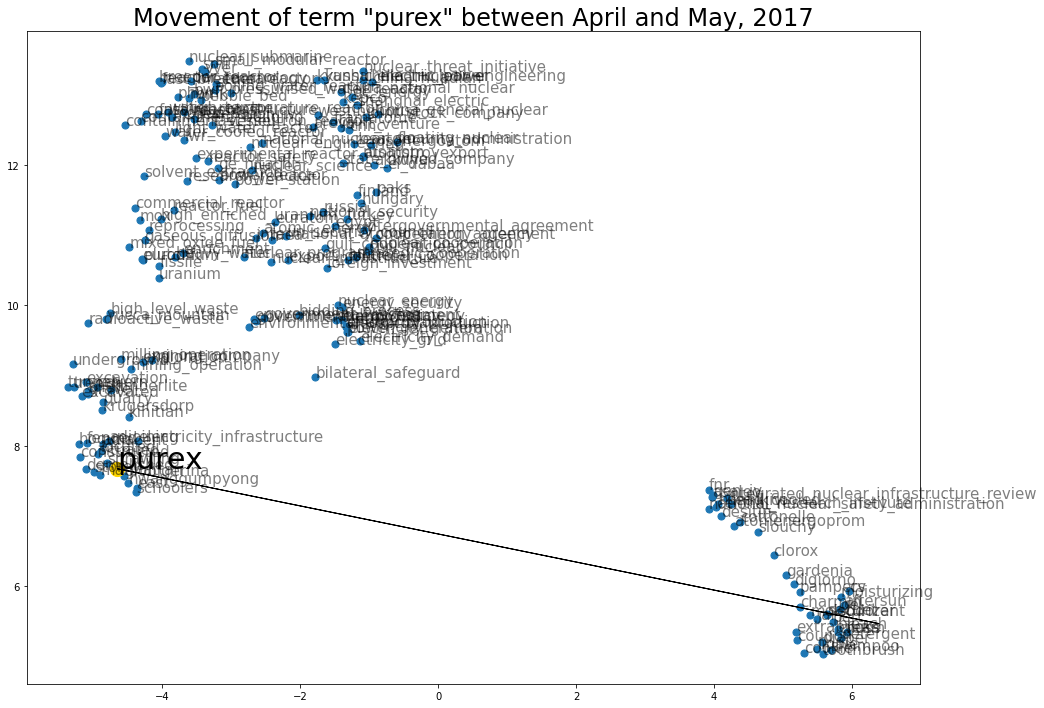}
    \caption{Scatterplot of UMAP representation of term ``purex'' between April and May in relation to other key terms.}
    \label{fig:base_scatterplot}
    \vspace{-1em}
\end{figure}

\changed{As mentioned, it is infeasible for users to examine millions of documents by hand to find significant events. Therefore, the baseline approach for identifying events from a corpus is to track the evolution of words over time through temporal word embeddings.} 
Temporal word embeddings align word vectors across time, capturing changes in usage over time through changes to the word vectors.
These embeddings enable several useful visualizations that allow users to pinpoint drastic word movement, aiding the discovery of events.

To demonstrate a potential workflow, the user can examine a heatmap of word embedding movement (\cref{fig:base_heatmap}) to gain a broad understanding.
A lack of movement demonstrates a consistent usage of a term between time slices, while greater movements mean sudden change.
The term ``purex'' experiences a jump in word vector position between April and May of 2017.
This corresponds to a significant change in the usage of the term ``purex'' during that time frame compared to the usual tendency for its representative word embedding to remain static.

A scatterplot of these word embedding movements (\cref{fig:base_scatterplot}) can highlight the shift from one context to another.
Initially, the context containing ``purex'' refers to the laundry detergent brand, as shown by its relatedness to terms describing consumer bathroom and laundry products (``coupon,'' ``shampoo,'' ``deodorant'').
However, there was a quick shift during this time period related to the ``Plutonium Uranium Extraction Plant'' (shown by its closeness to the terms ``underground'' and ``tunnel''), the quarry of which experienced a collapse during that time%
\footnote{See: \url{https://www.hanford.gov/page.cfm/purex}}.
Other terms can be analyzed in the same way to try and extract events of interest.


\changed{Though these methods may provide insight, they suffer from several shortcomings. First, they do not provide a broader narrative relative to the documents or a global topic space. Second, word embedding movements only give small piece of a bigger picture. While they may help illustrate differences across a single time step, users must manually inspect a series of time steps to get a full picture of a keywords evolution and how both it and the words around it move. This adds substantial burden to users, especially in the presence of many evolving keywords.
Finally, most events will not be as apparent as a word vector switching associations from cleaning products to nuclear energy, thus users need more guidance towards potential events. }

\section{Methodology}
\label{sec:method}

\begin{figure}
    \centering
    \includegraphics[trim={1cm 6cm 8cm 3.5cm},clip, scale=0.5]{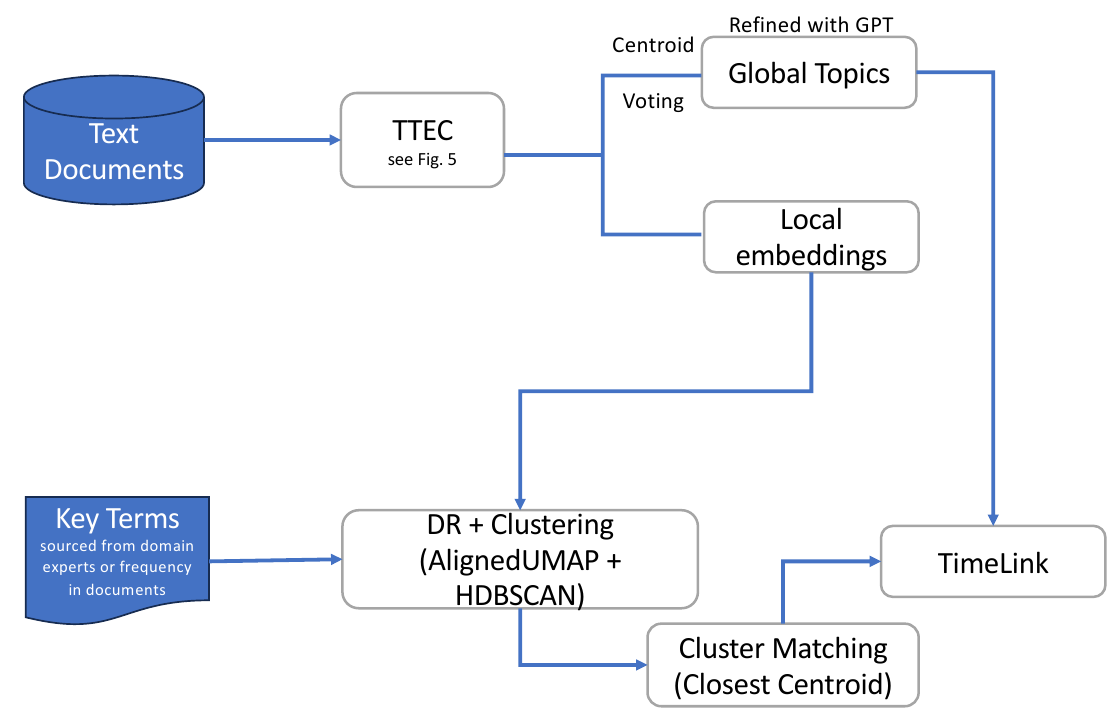}
    \caption{Architecture diagram}
    \label{fig:vis_pipeline}
    \vspace{-2em}
\end{figure}

\begin{figure*}
    \centering
    \includegraphics[width=0.8\textwidth]{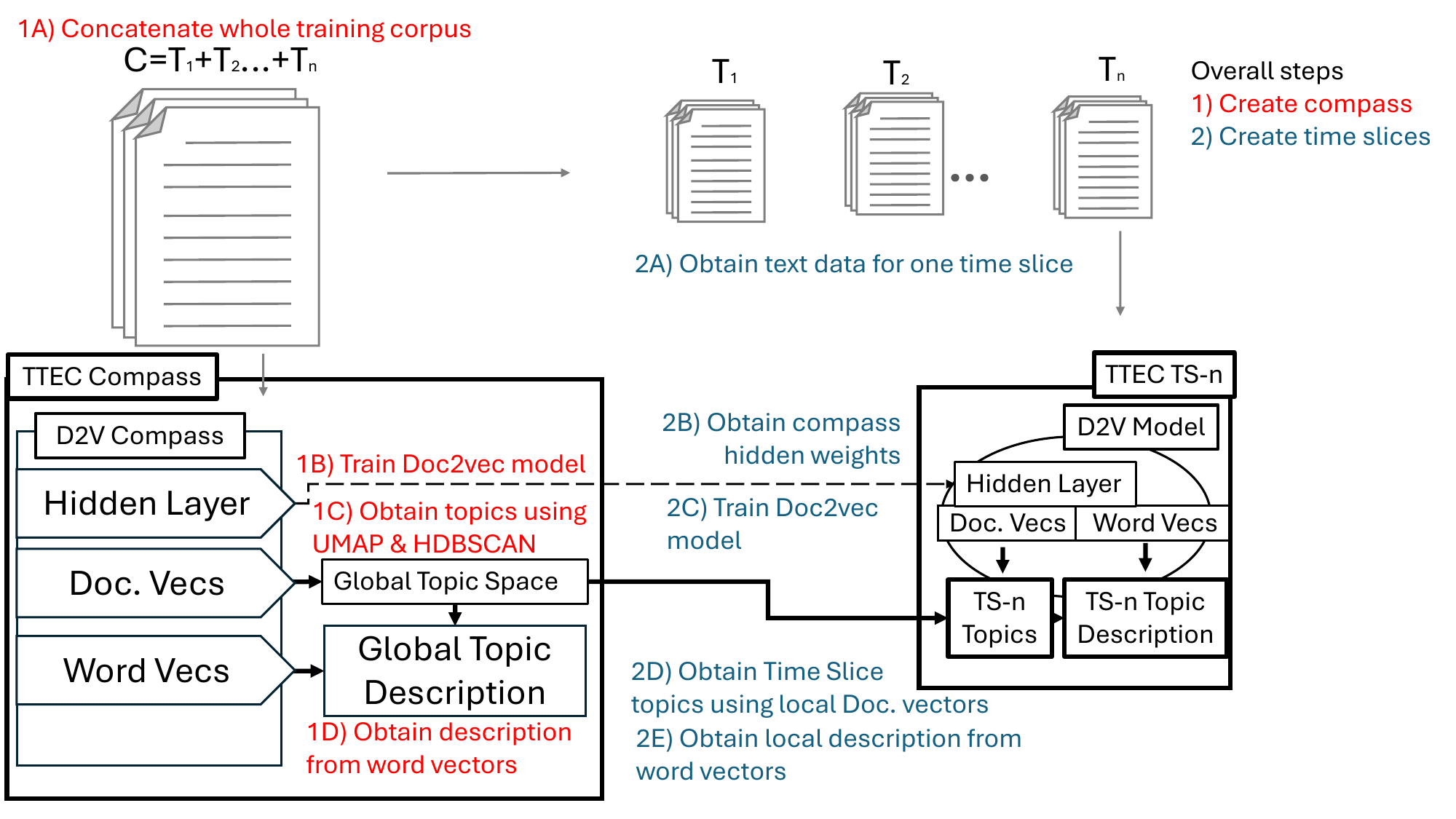}
    \vspace{-.5em}
    \caption{
    \changed{TTEC Architecture.
    Step (1) illustrates how the TTEC compass is trained through creating a TDEC Doc2vec compass (1A, 1B), and then creating a topic space based on the documents using UMAP and HDBSCAN (1C) and generating a topic description using global word vectors (1D).
    Step (2) looks at how individual time slices are trained.
    A TDEC Doc2vec time slice is trained (2C) using the local time slice corpus (2A) and compass hidden weights (2B).
    The local documents are then placed into the global topic space to create local topics (2D) and topic descriptions (2E).
    The local word vectors and topics will be used in TimeLink (Figures \ref{fig:vis_pipeline} and \ref{fig:sankeydash}).}
    }
    \label{fig:ttec_architecture}
    \vspace{-1em}
\end{figure*}

This section describes the methodology \changed{behind TimeLink, our VA system, and our two embedding methods:} Temporal Document Embeddings with a Compass (TDEC)\changed{, which enables us to compare temporal words and documents in the same embedding space,} and Temporal Topic Embeddings with a Compass (TTEC), \changed{which is a DTM method that creates global topic representations on top of temporal words and documents.
A general pipeline overview is shown in \cref{fig:vis_pipeline}.
Temporal Word Embedding methods can be used to capture the context of terms over time.
However, they lack the broader analytical context of documents.
TDEC addresses this by creating a compass-aligned Temporal Word and Document Embedding model for document context.
TTEC is a DTM method that extends TDEC by creating global topic representations out of the aligned documents and assigning those topics to aligned local word vectors.
The TTEC compass alignment and topic representations allows us to visualize the relationship between global topics and terms in a local cluster over time in TimeLink.
}
 

\subsection{Temporal Document Embedding with Compass (TDEC)}

TTEC aims to apply the compass alignment~\cite{bianchi20} approach to the creation of dynamic topics.
However, extending the compass metaphor to DTM requires first expanding the methodology to support document embeddings, thus the creation of TDEC.

\subsubsection{Compass Generation}

TDEC trains the aligning compass from the entire text corpus using the Doc2vec \cite{le14} method.
Functionally, this is identical to the TWEC compass creation process, except it substitutes Work2vec for Doc2vec.
As a result, global word and document vectors are created in addition to the hidden layer that will be used in time slice training.

\subsubsection{Time Slice Training}
\label{subsub:tdec_ts}

TDEC time slice training is similar to TWEC, except it uses the Doc2vec model as the time slice model instead of Word2vec.
First, the model initializes its vocabulary, using the contents of the time slice it represents.
Next, TDEC initializes the weights of the hidden layer in this Doc2vec model with the corresponding parts in the compass hidden layer.
For example, if a time slice model has ``labor'' in its vocabulary, then the weights that lead to the output for ``labor'' will be copied to the time slice from the compass.
\changed{The weights are then frozen and the model trained.}
Because each time slice shares the compass' hidden layer, this results in aligned word and document embeddings across time slices.

\subsection{Temporal Topic Embeddings with a Compass (TTEC)}
\label{sub:ttec}

After TDEC creates document embeddings with the compass alignment methodology, TTEC can further expand it to DTM.
The initial training process for this method is identical to that of TDEC, by creating a document-based compass and time slices.
Afterward, TTEC creates a global topic representation using UMAP and HDBSCAN that it then shares with each time slice.

\subsubsection{Compass Generation}

The creation of this compass first follows the TDEC method and can be made using either the PV-DM or the PV-DBOW model (Step 1A in Figure \ref{fig:ttec_architecture}).
TTEC then builds topics on top of the document embedding compass.
Topic generation (1B in Figure \ref{fig:ttec_architecture}) is a two-step process and follows the unsupervised topic generation process of combining UMAP and HDBSCAN.
First, TTEC performs UMAP \cite{mcinnes20} on the global document vectors, using cosine similarity as the closeness metric.
Afterward, it performs a hierarchical density-based clustering in HDBSCAN \cite{campello13hdbscan} \changed{to discover global topics}.
In the event of too many topics, TTEC repeatedly merges the smallest topic with the closest topic, as measured by Euclidean distance between topic centroids, until it reaches the desired number of topics.

\subsubsection{Time Slice Training}

During this step, local time slices obtain their local word, document, and topic representations.
\changed{A TDEC time slice is first trained to obtain aligned local document and word embeddings}
TTEC then performs an additional step to obtain local document topics.
Since the local and global word and document vectors all occupy the same embedding space, they can be placed into the global UMAP space.
Afterward, these newly placed points can be assigned a topic within the global UMAP space based on their proximity to the closest topics.

\subsubsection{Topic Descriptor Selection}

\begin{figure}
    \centering
    \includegraphics[trim={0 1cm 0 3cm},clip, width=\linewidth]{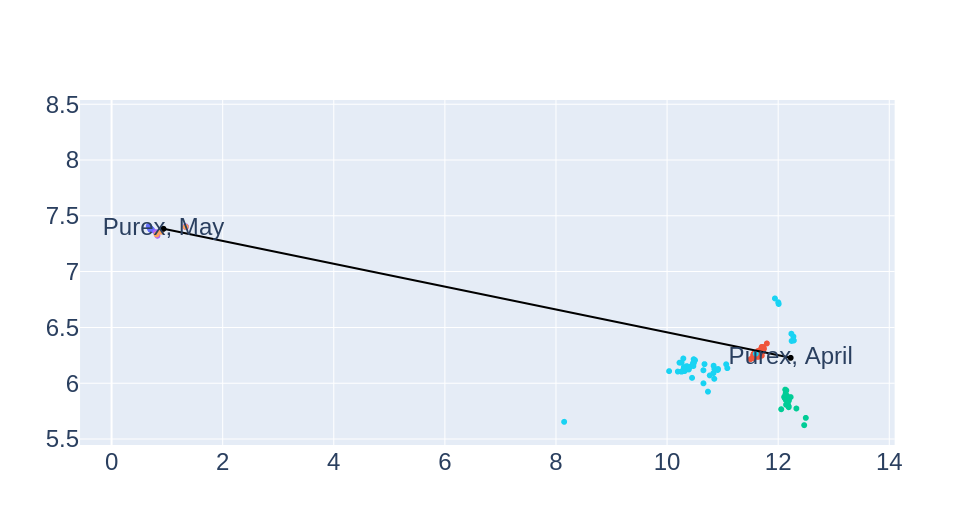}
   \vspace{-2em}
    \caption{Baseline Purex comparison with documents and DTM.
    In April, Purex is near topics related to health (aqua), nutrition (red), and home appliances (green).
    In May, Purex is near geology/radioactive material (purple), archeology (blue), and radioactive material/misc. (orange)
    }
    \label{fig:purex_topics}
    \vspace{-1em}
\end{figure}

Once TTEC generates the individual time slices, there is a question of how to describe topics in the time slices.
This was previously done in doc2vec-based topic modeling \cite{angelov20} by calculating the top \verb|n| most relevant words to a cluster.
It is also possible to use a class-based TF-IDF \cite{grootendorst22bertopic}. 
We present two methods by which these topic descriptors are selected:
centroid and voting.

The centroid method averages out all the document vectors in a local topic time slice to obtain a ``topic vector.''
The \verb|n| most similar words based on cosine similarity to this topic vector are then used to describe the topic at that particular time slice.
Obtaining the most similar words of a topic at every time slice allows for a temporal adjustment of topic descriptors, as terminology is added and phased out.
This method works well for a topic with points distributed in a Gaussian distribution, but does not work for atypical cluster shapes.

The voting method aims to take cluster shape into account.
For every document vector of a cluster in a given time slice, the top \verb|n| most similar words are found and kept track of.
The top \verb|n| words are then picked based on the overall results of this process.
This process results in topics influenced more by the densest parts of a topic space, which complements the density-based method of topic generation.
This paper chose to use the voting method due to this increased flexibility.

TTEC then allows an additional layer of analysis that goes beyond words by including document embeddings with topics as an overall abstraction (see~\cref{fig:purex_topics}).

\subsection{Visualizing Documents and Topics}

To illustrate the evolving context of keywords relative to global topics, we designed a visualization, TimeLink, that captures local topic structure and illustrates the movement of keywords between topics over time, while still maintaining context with global topics. The high-level process, shown in \cref{fig:vis_pipeline}, is as follows. First, we extract word embeddings from TTEC for the selected group of keywords. Then, for each time slice, we use dimension reduction to create low-dimensional representations of keyword embeddings and subsequently cluster them into local topics. With a set of local topics (i.e., clusters of keywords) for each time slice, we then match related clusters across time slices to illustrate global trends. Before visualizing these trends, we also generate global topic descriptors that we use to classify the local topics, to illustrate how the composition of keywords in global topics evolves over time. Finally, we create a Sankey diagram that encodes the clusters at each time slice (formed from individual keywords), the movement of keywords between clusters across time via connecting lines between clusters, and the global topic membership via color.  In the remainder of this section, we describe each step in detail.

\subsubsection{Dimension Reduction}
\changed{With the keyword embeddings extracted from TTEC, we first project the data using AlignedUMAP.}
AlignedUMAP creates aligned projections of a segmented dataset (e.g., by time) so that changes in projected position between segments result from data changes, rather than projection variance.
Thus, it fits our data where each dataset segment corresponds to time slice segments.

\begin{figure*}
    \centering
  \includegraphics[trim={0 6cm 0 0},clip,width=\textwidth, scale=0.20]{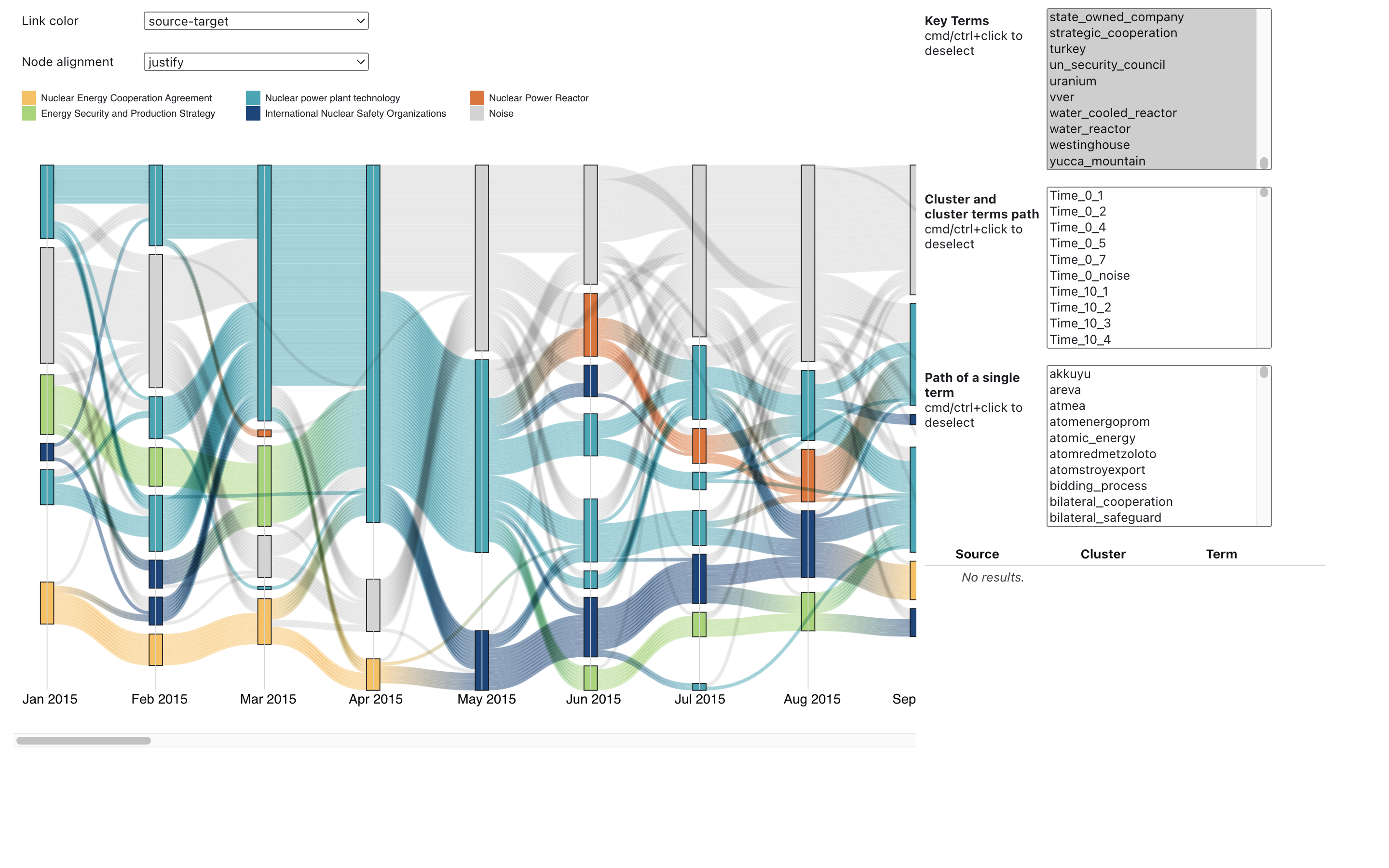}
  \caption{Dashboard with the default view of the first few time periods shown with a subset of the key terms selected.}
  \label{fig:sankeydash}
  \vspace{-1em}
\end{figure*}

\subsubsection{Clustering and Cluster Matching}


After projection, we cluster the keywords in each time period independently using HDBSCAN.
This helps capture the local topic structure in each time slice, relative to the keywords. 
\changed{However, this leads to disconnected clustering between time slices. Cluster 1 in one time slice might be called Cluster 3 in another, for instance.}
To address this, we examined two approaches: connecting two clusters based on their vocabulary and connecting two clusters based on the embedding space.
Any term not added to a cluster by HDBSCAN is considered noise \changed{and is treated as its own cluster}.

Connecting two clusters based on their vocabulary involves comparing one cluster in the current time period to all the clusters in the next time period. 
Clusters are connected based on the percentage of identical terms between them.
The cluster with the highest percentage in the next time period would be classified as the next time period's representation of the current cluster. 
The cluster with the most similarity typically has above 80\% similarity to the cluster it is being compared to and is a distinct classification. 
\changed{However, the method loses some validity when there is a split into two somewhat dissimilar clusters, or two clusters combine into a similar cluster in the next time period.}
These relationships should be considered in the visualization, but this may not be the best method since the only basis of the relationship between clusters is the terms.

The other approach connects clusters based on the embedding space. 
\changed{In it, we take the topic space composed of UMAP and HDBSCAN to find a low-dimensional representation of terms, global topics, and local topics across time slices.}
We use Euclidean distance \changed{within the reduced topic space} to evaluate the closeness of two clusters across time slices.
First, we find the centroid of the current cluster and the centroids of all clusters in the next time slice.
Then, we calculate the Euclidean distance between the centroid of the current cluster and the centroids of the clusters in the next time slice, selecting the cluster with the minimum distance \changed{to be the most similar cluster}.
\changed{This method introduces more variance but ends up being more reliable due to being based on a stable global embedding space.
A reliance on distance allows for new topics to emerge and not be connected to any past topics due to being far away.}
We choose to proceed with this approach to create the visualization because relationships are encoded in the global embedding space.

\subsubsection{Global Topic Descriptions}

Our methodology reveals clusters whose global structure is relatively stable, while the local structure provides meaningful insights into shifts in the global structure. 
Topic modeling adds explainability to the clusters and makes the global structures more meaningful.
Topics modeled using a global structure retain flexibility in interpreting how clusters change over consecutive time frames and longer periods; therefore, global topics are produced from TTEC, as sets of terms, and refined to be more descriptive using GPT-3.5 \cite{brown20}.

We match local clusters to global topics based on vocabulary. 
Each resulting topic returns the terms in the global corpus assigned to it and the probability of fitting in that global topic. 
The global topic with the most terms in common is assigned to that cluster. 

\begin{figure*}
  \includegraphics[trim={1cm 5cm 1cm 4cm},clip, scale=0.2, width=\textwidth]{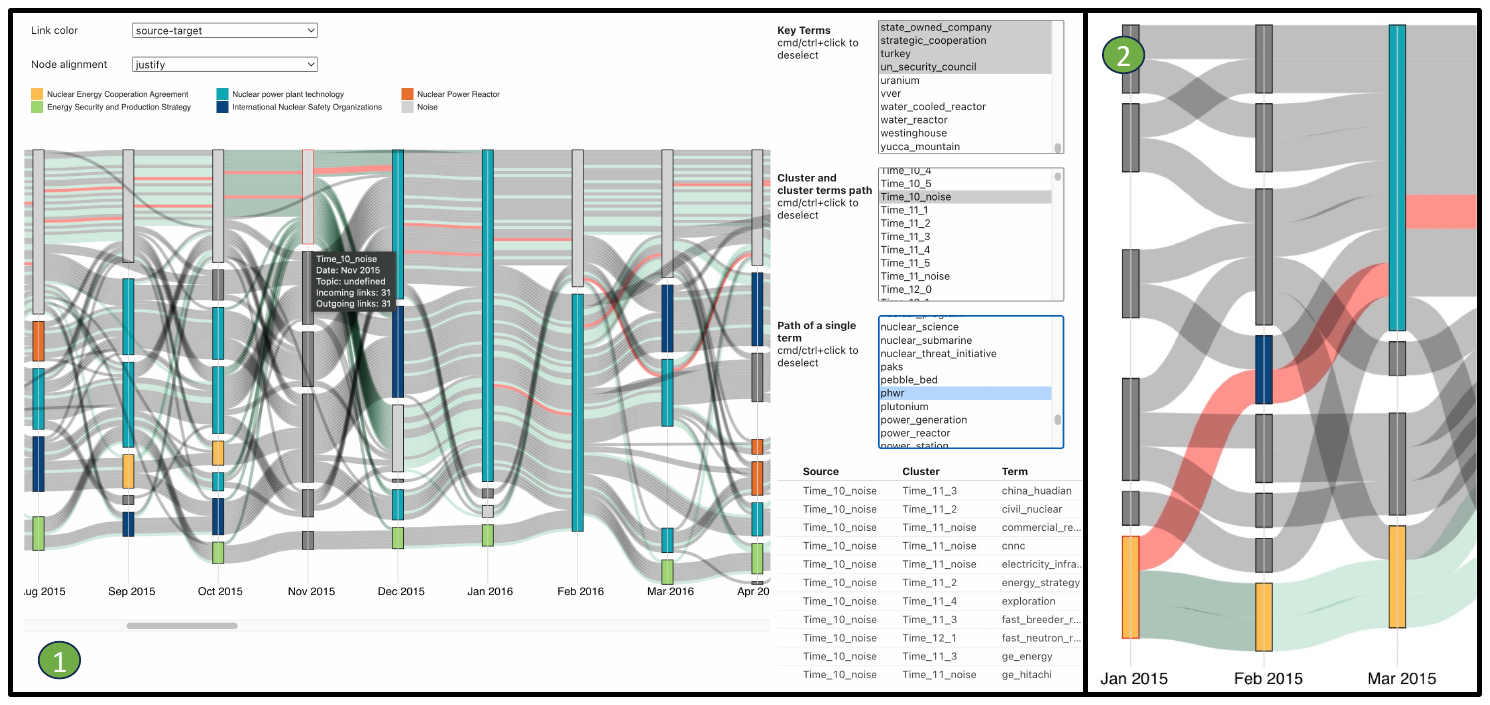}
  \caption{1. A view of the dashboard with a cluster selected. The dashboard now shows a table of terms in the selected cluster and details of nodes when a mouse hovers. 2. Zoomed in pathing of terms in cluster \textit{Time\_0\_1}   }
  \label{fig:sankeyclusterfilter}
  \vspace{-2em}
\end{figure*}

\subsubsection{TimeLink}
Using the matched clusters from the previous step, we visualize the evolution of topics and keywords with TimeLink. TimeLink uses a Sankey diagram to demonstrate the flow of keywords across local topics and the evolution of global topics. 

While Sankey diagrams typically illustrate energy flow between states, we employ them to visualize the flow of terms in clusters across time slices. In past applications to topic modeling, the data items in each state continually change. In contrast, TimeLink uses the same set of keywords in each state (time slice), creating links to connect clusters with shared keywords. Thus, the flow of the Sankey diagram emphasizes the movement of keywords rather than separation. 
This approach allows TimeLink to display movement using 5D or 10D vectors (which can show a higher variance) instead of 2D or 3D vectors (typically used for scatterplot visualizations).
This benefit comes with the limitation of finding an effective way to display the progression of time while preserving continuity and mapping to actual dates. 
However, TimeLink overcomes this limitation by placing more importance on word relations and movement. 

TimeLink's use of the Sankey diagram effectively shows the relevant global topics across all the time periods and their corresponding local cluster. 
The global topics are based on the placement of key terms and where they lie in the global topic embedding space alongside documents. 
The most significant challenge was showing how word embeddings statically move across time, and TimeLink overcomes this by providing a dynamic and interactive representation of movement across time.

\changed{The idea of TimeLink arose from the need to create a VA system that supports the interactivity of multi-domain data, data that concerns a variety of subjects like countries and local government agencies. 
With millions of documents, a variety of hierarchies and subjects are embedded within the data that a user can struggle to keep track of. 
There is also difficulty regarding how to visualize them over time. On a large scale, TimeLink was made to visualize multi-domain data to find answers to multi-domain problems. 
At its core,} TimeLink aims to illustrate interesting movement in temporal word embeddings. \changed{These problems and aims motivate TimeLink's} design \changed{which inspired} the following design goals:
\begin{enumerate}
    \item \textbf{Highlight time-based changes}: Users should, at an overview, be able to point out significant changes between multiple time periods quickly.
    \item \textbf{Find a word's relation to other words}: As the primary data for the visualization is word embeddings, users should be able to compare multiple key terms easily, e.g., if two key terms tend to stay close to each other in the embedding space across time or move in opposite directions during consecutive time periods. 
    \item \textbf{Support the discovery of insights rooted in real-world events}: How word embeddings develop over time is based on events and discourse in the corpus. The visualization should help users find interesting movements of key terms related to the document corpus and linked to a real-world event.

\end{enumerate}

The dashboard layout, shown in \cref{fig:sankeydash}, is designed for domain experts and layman users unfamiliar with the data. 
The center of the dashboard is the Sankey diagram itself. 
By default, TimeLink displays all keywords in the corpus. 
TimeLink encodes each keyword as a link. The rectangular nodes represent the local clusters in each time period, and each node's color is based on the global topic it is assigned to. 
TimeLink aligns nodes of the same time slice vertically such that all nodes in the same time period are stacked in a column, thus creating an ``axis'' that represents time. 
The links are displayed and colored based on the global topic of the node the link travels towards and travels from. 

Within the display window, by default, a user only sees 7-8 time periods at a time to aid with visual overload, but a user can scroll horizontally to display additional time slices.
A user can zoom out to see the full Sankey diagram with some precision. This design supports the first design goal of highlighting time-based changes. 
Users can easily spot significant changes, such as when a larger cluster splits in the following time period into smaller clusters with different global topic assignments (like the light blue cluster representing `Nuclear power plant technology' during May 2015 in \cref{fig:sankeydash}). 
This design also supports the second design goal, as users can see the word relations through a cluster context over a long period of time. 

The menus on the right include, from top to bottom, a text box where users select the words displayed on TimeLink, a text box where users can choose specific local clusters to highlight in the diagram, and a text box to select particular terms to highlight throughout the diagram.
The first text box allows users to adjust visualization size as needed.
In \cref{fig:sankeydash} and \cref{fig:sankeyclusterfilter}, the number of links displayed in TimeLink is filtered through the first text box.
As shown in \cref{fig:sankeyclusterfilter}.1, using the second text box to select a cluster, a few changes occur.
TimeLink colors any links and nodes that do not correspond to or contain terms from the selected cluster in gray.
The remaining links are colored to indicate their approximation to the selected cluster, outlined in red; links encoding keywords immediately traveling into or out of the selected cluster are colored dark teal, while all other links are colored a lighter shade of teal.
The nodes connecting the links throughout the diagram keep their original topic color.
The third text box only makes available the key terms selected from the first text box; when selected, the term's path is colored red.
This view also introduces a table on the right of the dashboard that shows the key terms in the cluster you selected. 
This can help a user focus on specific terms of interest, which can also be shown in the visualization by hovering over links that stand out to a user.
The tooltip shows the key term the link represents, the source of the term, and the target of the term.

These designs support the second and third design goals. 
Selecting and tracking the movement of desired key terms helps users focus on the immediate relationship of terms 
rather than having to create and maintain a mental picture while scrolling through the Sankey diagram. 
High-dimensional data centered around time can have many data points for many time periods. 
A flexible, detailed, and filtered view is crucial to ease users' cognitive load. 
A filtered view allows for faster discovery of insights that can reference real-world events.


\section{Case Study}
\label{sec:case}
This section presents two case studies to demonstrate the capabilities of the visualization system and illustrate different user workflows. We collaborated \changed{on the project} with \changed{two} domain experts in nuclear energy \changed{and natural language processing} who performed these analyses and offered suggestions for improved analysis in \cref{sub:qualFeed}. \changed{The development of the techniques was inspired by their need to streamline the identification of key points in time with maximal context that may have events of interest in large data corpuses.} Within these case studies, the users \changed{worked alone with only a simple introduction to the tool and} have access to documents \changed{that can be filtered} for relevant articles through the similarity metric provided by Word2vec and Doc2vec.

\subsection{Case Study 1: Cluster Analysis}
Using the text box "Cluster and cluster term paths" to select a cluster, it is simple to examine specific clusters and analyze how the relationship of terms \changed{(displayed by hovering over specific paths)} in that cluster develops over a period of time. \changed{For a domain specific corpus that is curated by a user, changes in cluster membership across time indicate a change in context and the potential occurrence of an event. Therefore, a user interested in understanding event trajectories across time may choose to display key terms of known interest and identify the movements of those terms between clusters across time. In this way, term movements between different topic clusters can be associated with documents such that events can be identified, providing more context when compared to simply tracking term movements in a temporal embedding model.}


\changed{As an example of events associated with changes in cluster membership,} the user \changed{populates the Sankey diagram in TimeLink with terms of interest and then} investigates cluster \textit{Time\_0\_1} and its future path (see \cref{fig:sankeyclusterfilter}.2) \changed{because a change was noticed.} 
At the initial time period, the terms in that cluster are \textit{rosatom}, \textit{cnnc}, and \textit{paks}. 
\textit{rosatom} represents the Rosatom State Nuclear Energy Corporation (Rosatom), a Russian state corporation located in Moscow that specializes in nuclear energy and high-tech products. 
\textit{cnnc} stands for the China National Nuclear Corporation (CNNC), a state-owned enterprise that oversees all aspects of China's civilian and military nuclear programs. 
\textit{paks} stands for the Paks Nuclear Power Plant (Paks) in Paks, Hungary. 
The global topic assigned to that cluster is the Nuclear Energy Cooperation Agreement, represented by the color yellow. 
The user selects \textit{Time\_0\_1} in the list of local clusters, highlighting the cluster at that time and the paths of those three terms throughout all time slices. 
At time 1, the following time period, \textit{paks} and \textit{rosatom} remain in the same cluster while \textit{cnnc} migrates to a different cluster in another global topic, International Safety Organizations (\textit{Time\_1\_5} colored dark blue).
This stability of the clustering of \textit{paks} and \textit{rosatom} in the two time periods makes the user believe that the two terms have a high contextual similarity. 
The user investigates this by checking some of the events identified in the documents containing those terms during this time period.  
They found that during that time period, the following example events were identified, proving their belief:

\begin{itemize}
    \item MOSCOW, October 1 (RIA Novosti) - A reactor at the Paks Nuclear Power Plant (NPP) in Hungary will begin operating on upgraded Russian nuclear fuel in early October, Russia's TVEL Fuel Company said Wednesday.
    \item Hungary and Russia signed an interstate agreement in January 2014 for Rosatom to build two new reactors to upgrade the Paks Nuclear Power Plant.
\end{itemize}

\begin{figure}

    \includegraphics[trim={1.5cm 1.5cm 0 3.5cm},clip, scale=0.28]{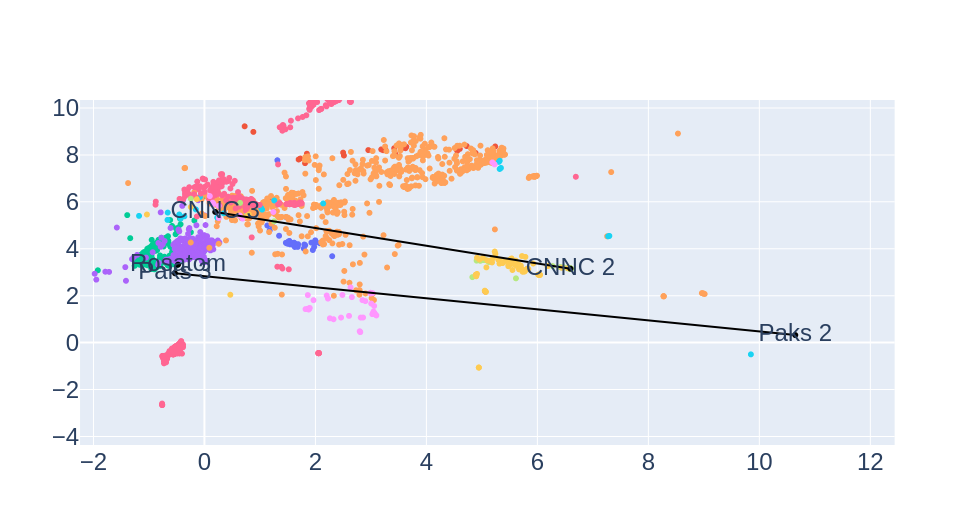}
  \caption{Zoomed in movement of CNNC and Paks towards Rosatom between \textit{Time\_2} and \textit{Time\_3} }
  \label{fig:cs1scatter}
  \vspace{-2em}
\end{figure}

At time 3, \textit{rosatom} migrates to the same cluster as \textit{cnnc} (\textit{Time\_3\_1} colored teal) while \textit{paks} moves to noise.
The user believes that the introduction of new articles in this time period strengthened the association between these two terms.
This relational movement is also seen in the document scatterplot \cref{fig:cs1scatter}, which highlights the topics of interest among hundreds.  
The user references the documents and finds that during this time, some news articles on international cooperation in civil nuclear energy strengthened the association of \textit{rosatom} and \textit{cnnc}:
\begin{itemize}
    \item CNNC : Rosatom may invite Chinese firm to nuclear plant construction
    \item Ukraine might cut gas imports from Russia thanks to joint deals with China.
\end{itemize}

The user continues to follow this pattern to find a time period when the terms rejoin the same cluster. 
This happens at time 11 when they enter cluster \textit{Time\_11\_2}, assigned to the global topic International Nuclear Safety Organizations. 
The user finds an explanation for this change in the documents; as news outlets discuss Rosatom’s and CNNC’s international collaborations, Paks returned to the discussion. 
Some of the  events that support this are: 

\begin{itemize}
    \item The project to build two new blocks of 1,200 megawatts each at the Paks nuclear power plant, awarded to Russia's Rosatom, has raised concerns that Hungary, a member of the European Union and NATO, was drifting back into Moscow's orbit.
    \item Paris (AFP) - French nuclear group Areva said that the state-owned China National Nuclear Corporation (CNNC) could take a minority stake in its capital under a draft deal signed in Beijing.
\end{itemize}

In summary, the user was able to analyze clusters, the terms they contain, and global topic cluster labels to gain relevant context into why the terms shift as they do. 
Because the global topics are rooted in a global embedding space containing the terms and the documents, they gave context to the user's exploration of cluster \textit{Time\_0\_1}.

\subsection{Case Study 2: Overview Analysis}
TimeLink allows users to examine how all the data moves over multiple time periods. 
The different encodings and filtering allow users to easily see a shift in the data and investigate its relation to a specific term. 

For this case, the user \changed{again populates the Sankey diagram in TimeLink with terms of interest and then} scrolls through TimeLink to look for significant changes and notices that a sizeable number of terms leave the noise cluster in \textit{Time\_10\_noise}, to join cluster \textit{Time\_11\_3} which gets the majority of its terms from \textit{Time\_10\_3} and \textit{Time\_10\_5}, shown in the third column in \cref{fig:sankeyclusterfilter} colored gray. 
\textit{Time\_10\_3} and \textit{Time\_10\_5} are assigned to the global topic, \textit{Nuclear Power Plant Technology}, and are colored teal. 

The user is curious about what terms left cluster \textit{Time\_10\_noise} to join \textit{Time\_11\_3} and their movement before this time. 
The user selects \textit{Time\_10\_noise} and \textit{Time\_11\_3} in the list of local clusters in the second text box on the right in \cref{fig:sankeyclusterfilter} labeled "Cluster and cluster term paths." 
The paths of all the terms in those clusters are then highlighted throughout the diagram, and a table containing the specific paths between time 10 and time 11 for each term in those clusters is laid out.

Since the user intends only to find the terms in both \textit{Time\_10\_noise} and \textit{Time\_11\_3}, they only need to scroll through the table to find the terms that occur twice and checkmark to highlight to refer to.
Those shared terms are \textit{china\_huadian}, \textit{fast\_breeder\_reactor}, \textit{ge\_energy, kepco}, \textit{kyushu\_electric\_power}, \textit{phwr}, \textit{rosatom}, and \textit{westinghouse}. 
To see the paths of these terms before time 10, the user selects them in the third text box labeled “Path of a single term” and the user notices that, although these terms are in \textit{cluster Time\_10\_noise}, the only terms that are consistently in a noise cluster over the three previous time periods are \textit{kyushu\_electric\_power} and \textit{phwr} (red paths in \cref{fig:sankeyclusterfilter}). 

\textit{kyushu\_electric\_power} refers to Kyūshū Electric Power Company (Kyushu Electric), a Japanese energy company, and \textit{phwr} is shorthand for a pressurized heavy-water reactor. 
The user hypothesizes that before time 11, \textit{kyushu\_electric\_power} and \textit{phwr} were not prevalent in the documents due to a lack of focus on nuclear energy. The two terms have contextual similarities due to their similar movements, so the user might theorize that Kyushu Electric may use a pressurized heavy-water reactor in some capacity. 
This encourages the user to further explore by referencing the document corpus to understand why these two terms were clustered as noise for so long and what brought them out of noise. 
The user finds by searching the document corpus by the keywords during this time period that Kyushu Electric is going to begin commercial operation of nuclear energy with the Sendai reactor:
\begin{itemize}
    \item Kyushu Electric reactor set to begin full commercial operation 
    \item Kyushu Electric begins commercial operation of Sendai reactor
\end{itemize}

Further research confirms the Sendai reactor is at the Sendai Nuclear Power Plant, which is run by Kyushu Electric and comprises two pressurized water reactors. 
Additionally, the user can hypothesize that the departure of \textit{kyushu\_electric\_power} and \textit{phwr} from being classified as noise might coincide with Kyushu Electric’s increase in nuclear energy production. 
Sampled events from documents containing these keywords during time 11 show unexpected events to the user:
\begin{itemize}
    \item Terror attack response centers for Kyushu Electric's restarted reactors announced
    \item Thousands take part in nuclear disaster drill near  Kyushu Genkai plant
\end{itemize}
The user now realizes that some significant events must have occurred between time 8 and time 11 for terror attack response centers and nuclear disaster drills to take place, which would indicate to the user that Kyushu Electric became relevant for a reason other than a rise in nuclear energy production.
The documents from time 9 reveal that a negative public sentiment was growing toward the company as it started to build a second nuclear reactor and an earthquake occurred nearby. 

The user then infers that the public scrutiny, evidenced by the documents in time 11 of the company’s conduct in the face of a natural disaster, brought Kyushu Electric out of being classified as noise. 
Referring back to the visualization, the user, after filtering the highlighted terms to only \textit{kyushu\_electric\_power} and \textit{phwr}, notices that the highlighted terms move to different clusters during time 13 (\textit{Time\_13\_noise} colored gray and \textit{Time\_13\_1} colored teal). 
The user would likely assume that the contextual similarity between the two terms weakened.
The user assumes that Kyushu Electric, an entity rather than a type of technology, was prevalent within the document corpus for reasons outside nuclear energy involvement.
This shift can also signal that a significant event occurred during this period. 
The user's hunch holds as documents support this train of thought; Kyushu Electric did start to get embroiled in more public scrutiny:
\begin{itemize}
    \item Kyushu Electric accused of 'cheap trick' in forgoing quake-proof center at Sendai plant
    \item EDITORIAL: Scrapped emergency plan puts Kyushu Electric's safety commitment in doubt
\end{itemize}
The terms return to the same cluster (\textit{Time\_15\_noise} colored gray), as noise in time 15, bringing the user back to their original assumption that the terms have high contextual similarity but are not mentioned much in relation to nuclear energy.
The documents indicate that events around Kyushu started to have less relation to Kyushu Electric. Events that specifically refer to Kyushu Electric mention the pressurized water reactor, but the events are not rooted in nuclear energy: 
\begin{itemize}
    \item Kyushu Electric Power: High court rejects residents' call to halt reactors in southwest Japan | 4-Traders
    \item Kyushu Electric Power: High court upholds ruling endorsing restart of Sendai reactor | 4-Traders
\end{itemize}

In summary, the broad overview of the visualization quickly allowed the user to see the movement of terms (together and apart) over multiple time periods and gain context on the changing inter-term relationship. 

\subsection{Qualitative Feedback} 
\label{sub:qualFeed}
\changed{In both case studies, the domain experts were able to meet their need of quickly identifying changes over time that point to the potential occurrence of an event. Compared to the baseline of navigating by hand through a large data corpus with a broad article domain that spans several years, domain experts used TimeLink as a guiding tool that quickly helped them to hone in on interesting events. TimeLink also gave them insight on how two events, identified through changes in cluster membership, are connected to each other by pointing to the causes and/or effects of the events.} 

During a user interview, a domain expert in nuclear energy who used TTEC and TimeLink considered the methods' structure and visualization highly applicable for seeking to understand logical and/or causal connections across trajectories of events that span multiple knowledge domains and extended periods of time. They further said that since the approach connects word and document embeddings to topics, it can aid in elucidating implicit connections. It is highly applicable to both historical research and monitoring of real-time streaming data. They suggested that another useful application of TimeLink is aiding hypothesis exploration because it offers a visual representation of how knowledge domains appear in reality when compared to users' expectations. The hierarchical aspect of the approach significantly reduces the need to perform manual searching. Notably, the techniques provided robust representations across time in the tested domains. 

The expert noted that use and functionality as a counterfactual tool may merit more future exploration.
In cases where sparse representation of data exists for a specific subtopic, or where significant noise is present in a dataset, it may be desirable to allow for human input, whereby the representation across time can be improved dynamically.

\section{Quantitative Results}
\label{sec:quant}

Having demonstrated the visual capabilities of temporal word embeddings aided by DTM, we next present an empirical analysis.
TTEC is ultimately a DTM method, and thus can be compared to other dynamic topic methods in terms of coherence and diversity of topics.

\subsection{Datasets}

TTEC was tested against the current state of the art in DTM using the Machine Learning Paper Corpus (MLPC), United Nations General Corpus (UN), and Nuclear Corpus (NC) datasets.
Preprocessing was standard across all three datasets, with fixed casing, removal of numbers and punctuation, and lemmatization.

The MLPC dataset \cite{rudolph18dynamic} is a collection of 17,772 machine learning papers from arxiv between 2007 and 2015 broken into 9 time slices.
The dataset is shared in a computer-readable format, so it had to be converted back to text data and turned into individual documents using the training and testing files.

The UN dataset \cite{baturo17un} is a collection of 7,507 speeches made during the United Nations General Debate between 1970 and 2015.
This paper tests on speeches made during the years 2006-2015, resulting in 10 time slices.
Since individual speeches might discuss opinions on a variety of topics across different paragraphs, paragraphs of each speech were separated into individual documents and documents with fewer than 20 characters were removed prior to pre-processing.


NC is the text corpus mentioned in \cref{sub:nuclear} (5,637,381 articles, 92 time slices), the purpose of which is to analyze the change in perception of topics related to nuclear energy.
This large dataset is the ultimate test of the scalability of DTM methods in this paper.
Given spatial and time constraints, we used a random sample of 10\% of the original data for the purposes of testing (563,739 articles, 92 time slices).
This random sample allows for a preservation of the distribution of articles monthly.

\subsection{Models}

We compared TTEC to the original DTM, called here Sequential Latent Dirichlet Allocation (S-LDA) \cite{blei06dynamic}.
We also compared it to the transformer-based Dynamic BERTopic (D-BTopic) \cite{grootendorst22bertopic} model.

\subsection{Testing Methodology}

TTEC was compared against Sequential Linear Dirichlet Allocation (SLDA) \cite{blei06dynamic} and Dynamic BERTopic (DBERT) \cite{grootendorst22bertopic}.
Comparison was made using Topic Coherence (TC) and Topic Diversity (TD) on the top 10 topic descriptors per topic.
TC was measured using a Normalized Pairwise Mutual Information (NPMI) \cite{bouma09npmi,roder15coherence} metric to see how likely words are to appear together and apart.
A coherent pair of topic words will have a NPMI closer to 1, 0 implies an independence of terms, while -1 means that a pair of words never appear together in the same document.
Coherence closer to one suggests that the topics are sensible.
TD is a measure of the proportion of unique topic words among all topics \cite{dieng20topic}.
TD closer to 1 implies distinct topics, while a TD closer to 0 implies redundancy within topics and their descriptions.
The aim is to have topics be coherent and diverse.
TD and TC are measured at every time slice with the average of 10, 20, 30, 40, and 50 topics and then averaged across all time slices.


\subsection{Tests}

\begin{table}[]
\begin{tabular}{lccccccccccc}
         & \multicolumn{2}{c}{\textbf{MLPC}} &  & \multicolumn{2}{c}{\textbf{UN}} &  & \multicolumn{2}{c}{\textbf{NC}} \\ \cline{2-3} \cline{5-6} \cline{8-9} \cline{11-12} 
         & TC              & TD              &  & TC             & TD             &  & TC             & TD             \\
S-LDA    & -.035           & .938            &  & .078           & .785           &  & DNF            & DNF            \\
D-BTopic & .006            & .957            &  & -.104           & .877          &  & .099           & .935           \\
TTEC     & -0.072          & .966            &  & -.307           & .985          &  & -.145          & .912             
\end{tabular}
\caption{Topic Coherence and Topic Diversity of different DTM methods across three datasets.
    }
    \label{tab:results}
    \vspace{-2em}
\end{table}

Looking at Table \ref{tab:results}, it is apparent that TTEC performs in certain types of data better than others.
With UN it was noticed that there was difficulty in capturing terms that would appear together in the same documents, with almost 70\% of pairwise comparisons resulting in no documents containing the two terms (and thus a coherence of -1).
This is partially due to the pre-processing breaking up text into smaller paragraphs that made capturing pairwise comparisons difficult.
There is no guarantee of terms existing in the text of the topic corpus as with tf-idf.
This was also the case with MLPC, although to a smaller extent, with 33\% of pairwise comparisons yielding no documents that shared the pair of terms.
With NC, there was an issue of rarer words and non-words that made it past the frequency filter to ruin the quality of some topics, while other topics depicted desirable topics of interest (domestic and international topics of nuclear energy production).

\section{Discussion}
\label{sec:disc}

\textbf{Performance and Scalability}
TTEC performs well with respect to the state of the art in DTM.
Notably, TTEC is trained using CPU resources and is parallelizable once the compass is trained due to a lack of reliance on previous time slices.
Similar parallelism was not possible to achieve with S-LDA, as shown by the DNF result.
If a GPU is available and analyzing the evolution of temporal word vectors is unnecessary, using BERTopic \cite{grootendorst22bertopic} could be beneficial for obtaining transformer-quality sentence embeddings.

As seen in \cref{fig:fullsankey}, in the presence of many time slices, TimeLink becomes longer and more difficult to view in full length. 
Alongside difficulty in viewing, the system's performance dwindles with a lag whenever a user scrolls further in time due to the large number of time slices and key terms. 
An immediate solution for performance, dependent on users, is to filter the amount of keywords used in TimeLink.
A more substantial solution is for users to increase the granularity of their time slices, resulting in fewer slices to visualize. However, this could result in aggregating away meaningful events. Another possible approach is to visually aggregate time slices with few meaningful shifts in topics, similar to DeltaRiver~\cite{gad2015themedelta}, to reduce visual clutter while illustrating important events. Additionally, in collaboration with domain experts, we found that experts only infrequently needed to view a large number of time slices simultaneously. Thus, in practice, scalability may not be a significant barrier. 

\textbf{Further Visual Design}
TimeLink enables users to understand how keywords shift in context and topic over time. However, additional user control in TimeLink could allow for more thorough analysis and understanding of the temporal changes. For example, as stated by the domain expert, the ability to incorporate user input would improve representation where there is sparse data or significant noise. We are pursuing further development of TimeLink into a broader event detection dashboard.

Initially, we explored the use of PCA and Aligned-UMAP plots to capture the changing keyword context between time periods by visualizing how points moved between time projections of time slices. However, we found that they were not stable enough to clearly illustrate keyword movements and changing clusters of keywords, thus the design of TimeLink. However, dimensionally reduced 2D scatterplots of individual time slices with both keywords and documents may help identify documents responsible for local topics. They may also aid in conceptualizing how closely related certain key terms are based on their location in the scatterplots. Thus, incorporating projections into a unified, linked dashboard with TimeLink may further improve analysis and should be explored in future work.

\textbf{Further Computational Evaluation}
The present quantitative evaluation does not have a temporal aspect to it.
Instead, atemporal measurements were taken at individual time slices and averaged to find an overall coherence and diversity score.
Future work would present measures that emphasize the temporal aspect akin to those presented in temporal word embedding literature\cite{yao18,carlo19}.





\section{Conclusion}

We introduced a novel method called TTEC for DTM with word embeddings and documents, and empirically evaluated TTEC performance against other DTM methods. It allows fine-grained insights gained through temporal word embeddings to be placed in the context of a global topic space and in the presence of documents that influence the word embeddings. Alongside TTEC, we presented TimeLink, a visualization system for temporal word embeddings and the movement of words within the context of global topic evolution, and validated the system through two case studies with domain experts. Together, TTEC and TimeLink address the challenges of combining words, documents, topics, and time
in an embedding space and visualizing that embedding space to reveal temporal changes. 

\acknowledgments{
This material is based upon work supported by the National Science Foundation under Grant \# 2127309 to the Computing Research Association for the CIFellows 2021 Project.}

\bibliographystyle{abbrv-doi-hyperref}

\bibliography{IEEEabrv,conference}

\end{document}